\documentclass[conference]{IEEEtran}
\IEEEoverridecommandlockouts
\usepackage[T1]{fontenc}
\usepackage[utf8]{inputenc}
\usepackage{cite}
\usepackage{amsmath,amssymb,amsfonts}
\usepackage{algorithmic}
\usepackage{graphicx}
\usepackage{textcomp}
\usepackage{xcolor}
\def\BibTeX{{\rm B\kern-.05em{\sc i\kern-.025em b}\kern-.08em
    T\kern-.1667em\lower.7ex\hbox{E}\kern-.125emX}}

\usepackage{amsthm}
\usepackage{amsmath}
\usepackage{amssymb}
\usepackage{amsfonts}
\usepackage{mathrsfs}
\usepackage{mathtools}
\usepackage{graphicx}
\usepackage{hyperref} \usepackage{cleveref} \usepackage{ifthen} \usepackage{xcolor} \usepackage{float} \usepackage{algorithm} \usepackage{algorithmic} \usepackage{hhline} \usepackage{url}

\newtheorem{prop}{Proposition}[section]

\newtheorem{thm}[prop]{Theorem}

\newenvironment{proof2}{\emph{Sketch of proof:}}{\hfill$\square$\vspace{1mm}}

\begin{document}

\newcommand{\sep}{\vspace{0.4cm}}

\newcommand{\sepno}{\vspace{0.4cm} \noindent}

\newcommand{\Rn}[1]
	{\ifthenelse{\equal{#1}{}}
		{\ensuremath{\mathbb{R}}}
		{\ensuremath{\mathbb{R}^{#1}}}}
  
\newcommand{\MAT}[1]	
	{\ensuremath{M_{#1}(\mathbb{R})}}		
		
\newcommand{\norm}[2]
	{\ifthenelse{\equal{#2}{}}
		{\ensuremath{\left\| #1 \right\|}}
		{\ensuremath{\left\| #1 \right\|_{#2}}}}

\newcommand{\argmax}{\ensuremath{\operatornamewithlimits{argmax}}}
		
\newcommand{\argmin}{\operatornamewithlimits{argmin}}

\newcommand{\rf}[1]{{\color{blue} #1}}
\newcommand{\ld}[1]{{\color{orange} #1}}
\newcommand{\warn}[1]{{\color{red} #1}}
\newcommand{\pat}[1]{{\color{magenta} #1}}
\newcommand{\kalou}[1]{{\color{green} #1}}

\title{Large scale Lasso with windowed active set\\ for convolutional spike sorting }

\author{\IEEEauthorblockN{
	Laurent Dragoni\textsuperscript{1}, Rémi Flamary\textsuperscript{2},  Karim Lounici\textsuperscript{3},  Patricia Reynaud-Bouret\textsuperscript{1}}
\IEEEauthorblockA{ 
	\textsuperscript{1} Université Côte d'Azur, CNRS, Laboratoire J.A. Dieudonné, 06108 Nice, France \\
	\textsuperscript{2} Université Côte d'Azur, CNRS, Laboratoire Lagrange, OCA, 06108 Nice, France \\
	\textsuperscript{3} École Polytechnique, Centre de Mathématiques Appliquées, 91128 Palaiseau, France \\
Email: \{laurent.dragoni, remi.flamary, patricia.reynaud-bouret\}@univ-cotedazur.fr, karim.lounici@polytechnique.edu}
}

\maketitle

\begin{abstract} 
Spike sorting is a fundamental preprocessing step in neuroscience that is central to access simultaneous but distinct neuronal activities and therefore to better understand the  animal or even human brain. But
numerical complexity limits studies that require processing
large scale datasets in terms of number of electrodes, neurons, spikes and length of the recorded
signals. We propose in this work a novel active set algorithm aimed at solving
the Lasso for a classical convolutional model. Our algorithm can be implemented
efficiently on parallel architecture and  has a linear
complexity w.r.t. the temporal dimensionality which ensures scaling and will
open the door to online spike sorting. We provide theoretical results about the
complexity of the algorithm and illustrate it in numerical experiments along
with results about the accuracy of the spike recovery and robustness to the
regularization parameter.  
\end{abstract}


\section{Introduction} \label{sec:intro}

The problem of spike sorting {consists in recovering the shape of action potentials and the time  activations (also called spikes) of individual neurons from the recordings of a population via an extracellular set of electrodes}. Since the shape of action potentials is usually characteristic of a given neuron and can be considered stationary,
it is possible to detect when an action potential occurs and to attribute it to a given neuron. This reconstruction of distinct simultaneous spike trains is a key preprocessing step to  evaluate phenomenons, such as firing rate coding or synchronizations  between neurons, that are presumably a sensible part of the neural code (see for instance \cite{albert2016,lambert2018,eytan2006synchro}).

As such it has been the focus of numerous
works in recent years
\cite{lewicki1998review,ekanadham2011recovery,ekanadham2014unified}.
{The estimation of the shapes is usually treated as a clustering problem, for shape estimation, followed by template matching to associate each action potential to a neuron having the closest shape \cite{pouzat2014}. While these methods have been
used in practice for a while, {especially in the case of tetrode recordings (meaning only 4 electrodes)}, they often require to perform manual
 pre and post-processing in addition to selecting several parameters, notably in presence of spike synchronization.}
 {As such the results of the procedure strongly depend on the person achieving the task \cite{wood2004variability, harris2000accuracy}. Moreover, these manual calibrations might not be possible in the near future due to the development of new acquisition methods and
the availability of larger datasets and recordings \cite{einevoll2012mea}.} Thus future spike sorting methods should depend on few parameters easy to select,
provide nice statistical properties and finally scale to large volume of data that will become the norm in the next few years: large number of electrodes {(up to 4000)},
large number of recorded neurons, {large number of spikes and synchronizations especially when the recordings takes place during epileptic seizures
\cite{neumann2017}}.

\begin{figure}		\centering
		\includegraphics[width=\columnwidth]{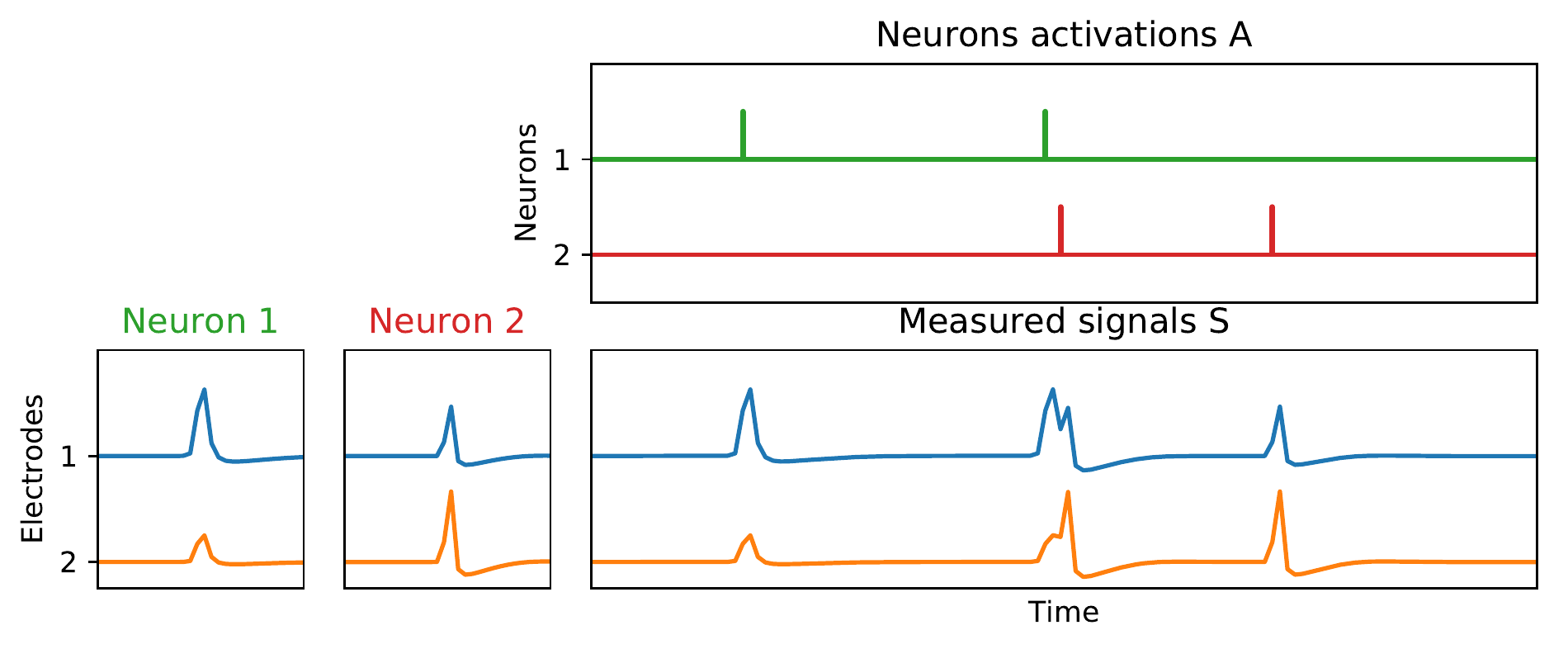}\vspace{-5mm}
		\caption{Illustration of the convolutive model with $d=2$ electrodes and $k=2$ neurons. Each activation of the neurons generates the associated shapes on both electrodes. Note the unusual shape of the action potential at the center resulting from an additive superposition of shapes which models synchronization of both neurons.		In the linear model \eqref{eq:vectorized_model}, such events are described as overlaps between two columns of the matrix $H$.}
		\label{fig:convolutive-model}
\end{figure}

\paragraph*{Convolutional model}
{{Apart from the classical clustering methods,}} one particular method relies on convex
optimization \cite{ekanadham2011recovery} and uses a convolutional model for  the recorded signals on $d$ electrodes.
 {To describe it, let us use the following notation: upper case letters correspond to matrices and lower case
letter to vectors, $*$ stands for the convolution operator along time. The model is written as follows:}
\begin{equation} \label{eq:convolutive_model}
	S = \sum_{r=1}^{k} W_r * A_r + N,
\end{equation}
where $S \in\mathbb{R} ^{d \times n}$ is the observed signal on $d$ electrodes
and $n$ temporal samples,  $W_r \in \Rn{d \times t}$ is the matrix containing
the shapes of the action potential of neuron $r$ on every {electrodes} {(all the shapes have a temporal spread of $t$)} and  $A_r \in
\Rn{1 \times n}$ is a sparse signal {called activation in the sequel} {(more precisely, the time activations of neuron $r$ corresponds to the indices of the non zero coefficients of $A_r$)}
and $N \in \Rn{d \times n}$ is a noise matrix. An illustration of the model is
provided in Figure \ref{fig:convolutive-model}.
This model has been introduced
for modeling speech signals in \cite{smaragdis2007} and has been  used {more recently} for
 spike sorting in \cite{ekanadham2011recovery,ekanadham2014unified}.   The main advantage is that this model explicitly takes into account synchronization as an additive superposition of shapes,
 which cannot be done with
simpler clustering based methods \cite{lewicki1998review}.

The
model \eqref{eq:convolutive_model} is usually estimated by alternative
optimization \emph{w.r.t.} the activations $A_r$ and the shapes $W_r$. But the
more computationally intensive update is clearly $A_r$ since the number of
variables in $A_r$ is proportional to $n$.
{Indeed, to give an order of magnitude, records last typically 30 minutes to hours, which means with the classical time resolution that $n\simeq 10^8$, whereas the number of electrodes $d$ range typically from 4 to 4000, the number of neurons from 1 to 1000  and the temporal spread of a shape  is of the order of the ms, hence $t$ lies between between 30 and 150 samples.}

This is why we focus on the problem of estimating $A_r$ when the shapes
$W_r$ are known.  In this case the estimation of the activations $A_r$ can be
done using the basis pursuit estimator as proposed in \cite{ekanadham2011recovery}. This
estimator, that is equivalent to the Lasso estimator, has several nice statistical properties such as the ability to
accurately recover the support, \emph{i.e.} activation times and active neurons, 
under {appropriate}
conditions \cite{BRT09,bunea2008,lounici2008}. 
The main problem with this estimator in this case is the
computational complexity of estimating the $A_r$ on large signals (large $n$).
Indeed the number of variables in the optimization problem grows linearly with
the length of the recorded signal. 
Several works have investigated  similar optimization problems 
in \cite{jas2017,la2018multivariate} but to the best of our knowledge, our
approach is the first to really exploit the structure of the problem to probably attain linear complexity in $n$.

\paragraph*{Contributions}

We propose a novel active set
algorithm working on temporal sliding windows that allows to estimate the
positions and magnitude of the sparse activations in model
\ref{eq:convolutive_model}. Our method relies on simple operations such as
convolution and can be easily implemented on parallel GPU architecture
which ensures a nice scaling \emph{w.r.t.} the dimensionality of the problem. We also prove for a simple yet realistic scenario on neurons activations that our active sets method will need to solve most of the time low-dimensional problems and even compute their average dimension. This last
result ensures that the complexity of solving the optimization problem will stay
linear \emph{in} the number of temporal samples $n$. Finally we perform numerical experiments that confirm the linearity of the algorithm in practice and
illustrate the robustness of the Lasso in this convolutional model for spike sorting.

\section{Optimization problem and large scale windowed active set algorithm}
\label{sec:model_and_activeset_algorithm}

\subsection{Large scale optimization problem for spike sorting}
\label{subsec:math_model}

{
Due to the linearity of the convolution operator, we can reformulate
\eqref{eq:convolutive_model} as a linear model. Vectorization of $S$ (resp: $N$)
as $y \in \Rn{dn}$ (resp: $\eta \in \Rn{dn}$), lead to the following form:
\begin{equation} \label{eq:vectorized_model}
	y = Ha + \eta ,
  \end{equation}
  where $a \in \Rn{kn}$ is the activation vector to be estimated, and $H =
  \MAT{dn,kn}$ is a block Toeplitz matrix, coding for the convolution between
  the shapes of the neurons and the activation vector. More precisely the blocks
  of the matrix $H_{p,r}$ are Toeplitz matrices of size $n \times n$, coding for
  the convolution between the shape of neuron $r$ on electrode $p$ and the
  activations of neuron $r$. 
  Matrix $H$ is very sparse and structure{d} with a sparsity ratio $\approx
  \frac{t}{n}$ with $t\ll n$. 
  }

{
Estimating $a$ when the number of neurons is bigger than the number of electrodes
would be impossible without additional structural assumptions. But biological
evidences show that neurons tend to spike rarely {(typically the number of non zero coefficients in $A_r$ should be of the order $10^5$ when $n\simeq 10^8$)} therefore the number of indices
to be activated in $a$ should remain small relative to $kn$. Since $a$ is a
sparse vector, we can use an estimator promoting sparsity,
such as the well-known Lasso estimator equivalent to what is proposed in \cite{ekanadham2011recovery} :
\begin{equation} \label{eq:lasso_problem}
    \hat{a} = \argmin_{a  \in \Rn{kn}} \frac{1}{2} \norm{y - Ha}{2}^2 + \lambda \norm{a}{1},
\end{equation}
where $\lambda>0$ is the only tuning parameter of the method that depends on the signal to noise ratio.

Standard approaches to solve problem \eqref{eq:lasso_problem} include well known
algorithms such as LARS \cite{efron2004least}, coordinate descent
\cite{wu2008coordinate} and more recently proximal gradient descent
\cite{combettes2011proximal,parikh2014proximal,beck2009fast}. When the solution
is known to be very sparse, active set approaches that iteratively add variables
and solve only low-dimensional Lasso problems have been known to provide fast solvers
\cite{boisbunon2014}. Other accelerations such as dual acceleration combined with screening and active set also work very well in practice
\cite{massias2017safe,massias2018celer}. 

In this work we propose a novel active set algorithm
that takes into account the structure of the problem.
The main ideas behind the optimization procedure
  are to i) use this sparsity to solve several small scale independent
  optimization problems and ii) to use the convolutional structure of $H$ to
  speedup computations and limit memory use.
}

\subsection{Active set for estimating the activations}
\label{subsec:basic_activeset}

{
The essential idea behind the active set (AS) algorithm is to solve only small Lasso problems \eqref{eq:lasso_problem}
by working only on a small subset of indices that are considered meaningful. The
subset of indices updated at each iteration is the active set and is denoted as
$J$ in the following. 
A crucial property of the solution of the Lasso problem comes from the following  \textit{KKT conditions} (see \cite{bach2011convex} for more detail).
\begin{prop}
	Let $a^*$ be a solution of \eqref{eq:lasso_problem} and write $H_j$ as the $j$-th column of $H$, where $1 \leq j \leq kn$. 	\begin{equation} \label{eq:kkt_condition}
	\forall j, \quad \text{ if } \quad	| H_j^T(y - Ha^*) | < \lambda,\quad \text{then } \quad a_j^* = 0.
	\end{equation}\end{prop}\vspace{-1mm}
The property above gives a good test to determine if a variable $a_i$ set at $0$ should
remain at $0$. Thus an idea would be to initialize the vector $a$ at $0$ and add
all the variables that violate this condition iteratively until all the remaining zero variables satisfy the condition. This leads to the AS algorithm
reported in Alg. \ref{algo:Lasso_resolution}. 
 Starting from a null
vector $a$ and an empty active set $J$, we add to $J$ the index $j \in J^c$
corresponding to the largest violation of the KKT condition $| H_j^T(y - Ha) |$.
Then we update our estimation of $a$ by computing the Lasso
solution for the problem $(H_J, y)$, where $H_J$ is the submatrix of $H$
obtained by keeping the columns $H_j$ of $H$, for $j \in J$. We repeat this step
until \eqref{eq:kkt_condition} is satisfied for all $j\in \{1,\dots,nk\}\backslash {J}$.
Overall the active set strategy will need to solve much smaller problems than
the original one when the solution is sparse but it requires access to
an efficient Lasso solver. }

\begin{algorithm}[t]
	\small
	\caption{Active set algorithm \label{algo:Lasso_resolution}}
	\label{algo:generic_active_set}
	\begin{algorithmic}[1]
		\REQUIRE $y,H,\lambda>0,\epsilon>0$
		\STATE $J\leftarrow \emptyset$, $a\leftarrow \mathbf{0}$
		\REPEAT
			\STATE $g \leftarrow |H^T (y-Ha)|$ \label{algo:correlation_computation}
			\STATE $j \leftarrow \argmax_{l \in J^c} g_l$\label{algo:max}
			\STATE $J \leftarrow J \cup \{j\}$\label{algo:add}
		\STATE $a_J \leftarrow$ Solve Lasso \eqref{eq:lasso_problem} for
		sub-problem $(H_J, y)$  \label{algo:sublass_solve}
		\UNTIL $g_j < \lambda + \epsilon$
		\RETURN $a$, $J$
	\end{algorithmic}
\end{algorithm}

\subsection{Optimization of the active set algorithm}
\label{sec:active_set_optimizations}

{
The active set algorithm described above is efficient {but may be further
optimized using the structure of the problem}. In the following, we  present optimization for the most
expensive steps in \Cref{algo:generic_active_set}: the computation of the KKT
condition (line~\ref{algo:correlation_computation}-\ref{algo:max}) and the Lasso estimation on
$J$ (line~\ref{algo:sublass_solve}).}

\paragraph{Memory and linear operator optimization}
\label{paragraph:memory_and_correlation_optimization}

{A first optimization that greatly reduces the time and space complexity is to
use  convolution to compute the KKT violation of line
\ref{algo:correlation_computation}. Note that $Ha$ can be computed with
$O(td|J|)$ using the sparsity of $a$ and the correlation $H^T (y-Ha)$ can be
computed with $O(nkdt)$ operations that is the main computational bottleneck.
The sub-problem in line \ref{algo:sublass_solve} is of size $|J|$ but
$H_J$ is actually very
sparse and has only $td|J|$ non-zero lines.  
}

\paragraph{Groups of activations}
\label{paragraph:groups_of_activations}

We can exploit the convolutional structure of our problem  to reduce the
cost of the Lasso step (line~\ref{algo:Lasso_resolution} in
\Cref{algo:generic_active_set}). Action potentials are short time localized
events, of length $t$ in our model. Then finding a new activation only affects a
portion of size $t$ in the signal. Therefore the only coordinates in $a$ that
are updated by the Lasso  are the ones in a temporal window of width $t$
around the {new activation}. Instead of updating $a$ on the whole active set, we only need to consider the indices {$i$} in $J$ that overlap with the new
activated sample {$j$, in the sense that $|i-j|\leq t$}. 

This means that the true complexity of the AS update actually depends only on how big is the set of indices that need to be updated each time.
{An order of magnitude of the  size of these overlapping groups (for short, overlaps) is computed in the next section, under plausible biological assumptions.}
{If this size does not depend on $n$,} this allows us to greatly decrease the size of the Lasso estimation
(line~\ref{algo:Lasso_resolution}). Note that a  similar approach was proposed in
\cite{boisbunon2014} for 2D convolution but required the use of a connected
component algorithm \cite{hopcroft1973algorithm} that is not necessary in 1D
where groups can be updated more efficiently.

{
\paragraph{Sliding window active set}
\label{paragraph:sliding_window} We now discuss the core of our approach. First note that when using the optimization above, the
computational bottleneck of the AS is KKT violation computation and maximum in lines {3 and 4}
since it requires to compute the  $O(nkdt)$ correlation in line
\ref{algo:correlation_computation} for each AS iteration. 
Since the number of iteration in the
AS is $O(kn)$ this leads to a quadratic complexity in $n$ which limit its
application on very long signals.

But finding the maximum along the whole signal is actually not
necessary for convergence of AS as long as the KKT conditions are verified at
the end. This is why we propose to compute KKT violation (line 3) and find the maximum {(line 4)} only in a temporal window of
size $w>t$ in practice. We start
with a window at the beginning of the signal and perform the active set. When
there is no violation of the KKT on the window, two possible things occur. If
there is an overlap strictly inside the window, then we can move the window to
the right since this group is independent from the rest of the signal and its
KKT will never be violated in the future. If the overlap goes beyond the window, then the size of the window is extended until the group is fully
contained in it and the window can be translated again to the right. What makes this
new approach very interesting in practice is that now the KKT violation in line
\ref{algo:correlation_computation} requires $O(wkdt)$ operations, that is independent on $n$.
This means that if the resolution of the sub-Lasso in line
\ref{algo:sublass_solve} does not depend on $n$, then the whole AS algorithm complexity becomes 
$O(n)$ instead of $O(n^2)$. 
}

\subsection{Size of the overlaps}
\label{subsec:overlaps}

{As we have seen before, the complexity depends strongly on the size of an overlap, meaning the larger set of time activations $\chi$ such that 
\begin{equation*}
    \chi=\{j \mbox{ such that } a_j\neq 0 \mbox{ and } \exists \ell \in \chi, 0<|j-\ell|\leq t\}.
\end{equation*}
Often Poisson processes are used to model time activations of neurons in neuroscience \cite{tuleau2014}. Of course, these processes are in continuous time, whereas our record is in discrete time. But the discretization is so thin with respect to the firing rate of the neurons that the difference is negligible. Moreover the discretization process usually only discards spikes \cite{tuleau2014} and can only diminishes the size of an overlap.}

\begin{thm} \label{thm:overlaps}
	Assume that the activations of the neurons in \ref{eq:convolutive_model} are drawn from independent Poisson point processes, that is
	\begin{equation}
		\forall 1 \leq r \leq k, \quad A_r \sim PPP(\mu_r),
	\end{equation}
	where $\mu_r > 0$ is the spiking intensity of neuron $r$. Let $\mu= \mu_1 + \dots + \mu_k$. Then the mean size of an overlap is smaller than $\mu t e^{\mu t}$. 
\end{thm}

\begin{proof2}
	{Consider $A$ the union of the $A_r$. $A$ is a Poisson point process of intensity $\mu$. The gaps between two activations 	obey an exponential distribution with mean $1/\mu$ and are independent, then we can compute the probability that the length of an overlap is greater that $mt$, with $m \in \mathbb{N}$}.
\end{proof2}

\Cref{thm:overlaps} gives an average size of overlap independent from $n$. Despite the exponential term, the small spiking intensity $\mu$ make this size reasonable in practice. For instance, for a sampling frequency $F=30$kHz, a spiking activity of $30 $Hz ($\mu=10^{-3}$), and spikes of length approximately $5.10^-{3}$s ($t=150$), we have an average number of activation in the overlap smaller than $7$ for $k=10$ and $60$ for $k=20$. Those values are in practice very small compared to $n=10^8$ corresponding to roughly an hour of recording. Note that when the number of electrodes becomes large we can extend this result to spatio-temporal overlap, leading to possible decrease in their size.

\section{Numerical experiments}
\label{sec:numerical_experiments}

{In this section we illustrate the performance and computational complexity
of our solver. For solving the Lasso, for both the full model and the AS
approaches we use the accelerated proximal gradient FISTA \cite{beck2009fast}
that is implemented in parallel in the SPAMS toolbox \cite{mairal2014spams}. 
All experiments were performed of a simple notebook having 8GB memory and an
Intel(R) Core(TM) i7-4810MQ CPU @ 2.80GHz. The
code will be made available online upon publication.}

\subsection{Computational complexity}
\label{sec:perf_and_complexity}

{
{In order to evaluate the computational complexity of our method we create a
dataset from simulated data.
We simulate realistic action potential shapes using the well-known
Hodgkin-Huxley model \cite{hodgkin1952} with implementation from \cite{pouzathuxley}, and generate matrix of activation $A_r$ as discrete point Poisson processes of intensity $\mu=10Hz$.  
We want to see the impact of the different optimization procedures discussed in
\ref{sec:active_set_optimizations}. First we solve the full Lasso  that requires
to pre-compute matrix $H$ which limits the size to the available memory (Global
Solver), we also report the performance of a naive AS that also use the $H$
matrix to see the effect of AS on sparse solutions. Finally we implement
the activation groups speedup (AS with
groups \ref{paragraph:groups_of_activations}) and the sliding window AS discussed in
\ref{paragraph:sliding_window}.}
Using $5$ neurons and $4$ electrodes, we
generate signals of various lengths (that is $n$), in the noiseless scenario.

\begin{figure}			\includegraphics[width=230pt]{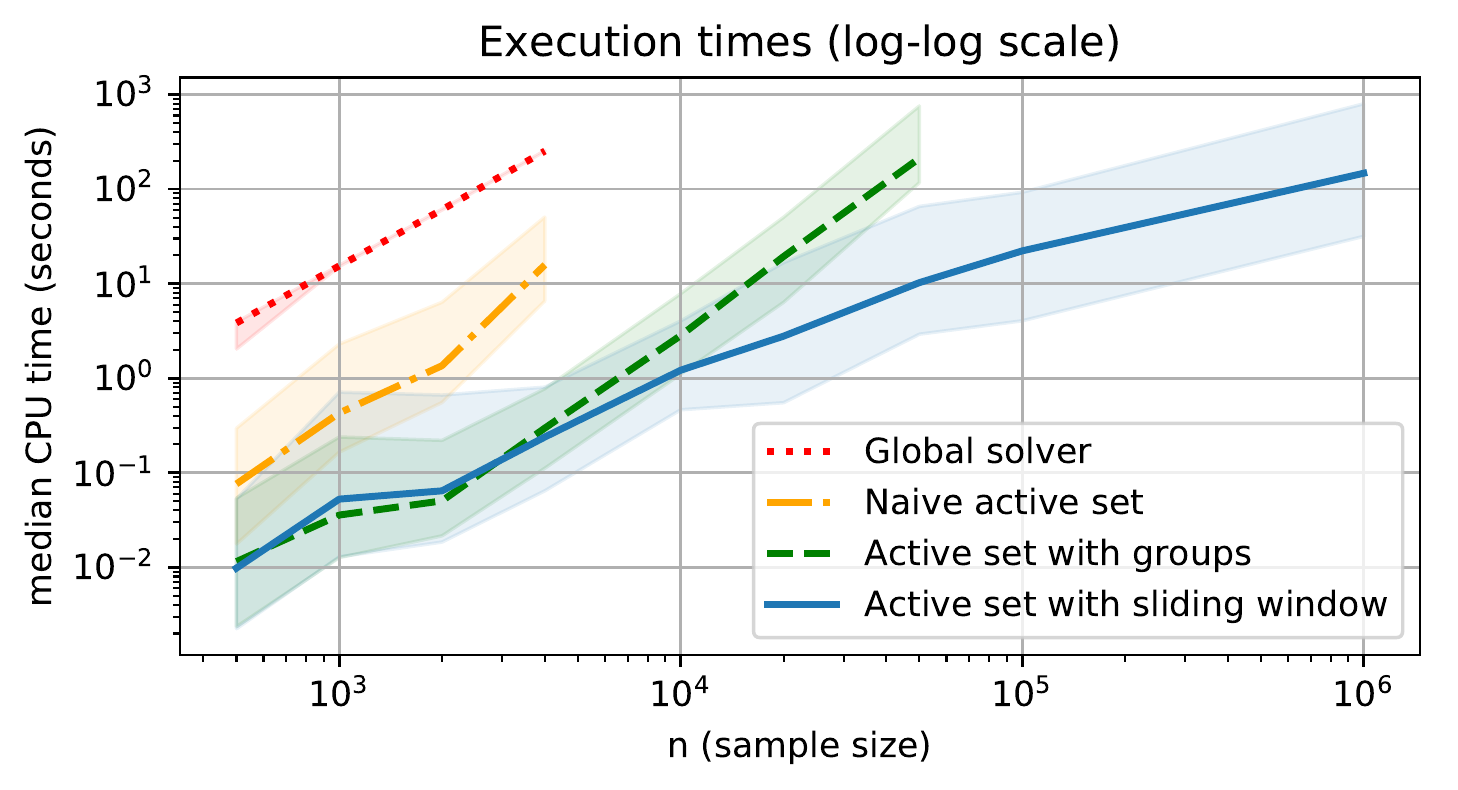}\vspace{-3mm}
		\caption{Comparison of the various algorithms in terms of execution times as functions of $n$. Results are averaged over 20 simulations.}
		\label{fig:exec_times}
\end{figure}

On \Cref{fig:exec_times}, we can visualize
the impact of the optimizations of the active set that we described in
\cref{sec:active_set_optimizations}.
The global solver and the naive AS can only handle small values of $n$,
as they are both memory intensive but the AS is one order of magnitude more
efficient. It shows that the execution
time of the active set with sliding window growths linearly with $n$ whereas the
AS on groups grows quadratically as discussed in the previous section.
}

\subsection{Influence of the noise and the regularization parameter}
\begin{figure}			\includegraphics[width=240pt]{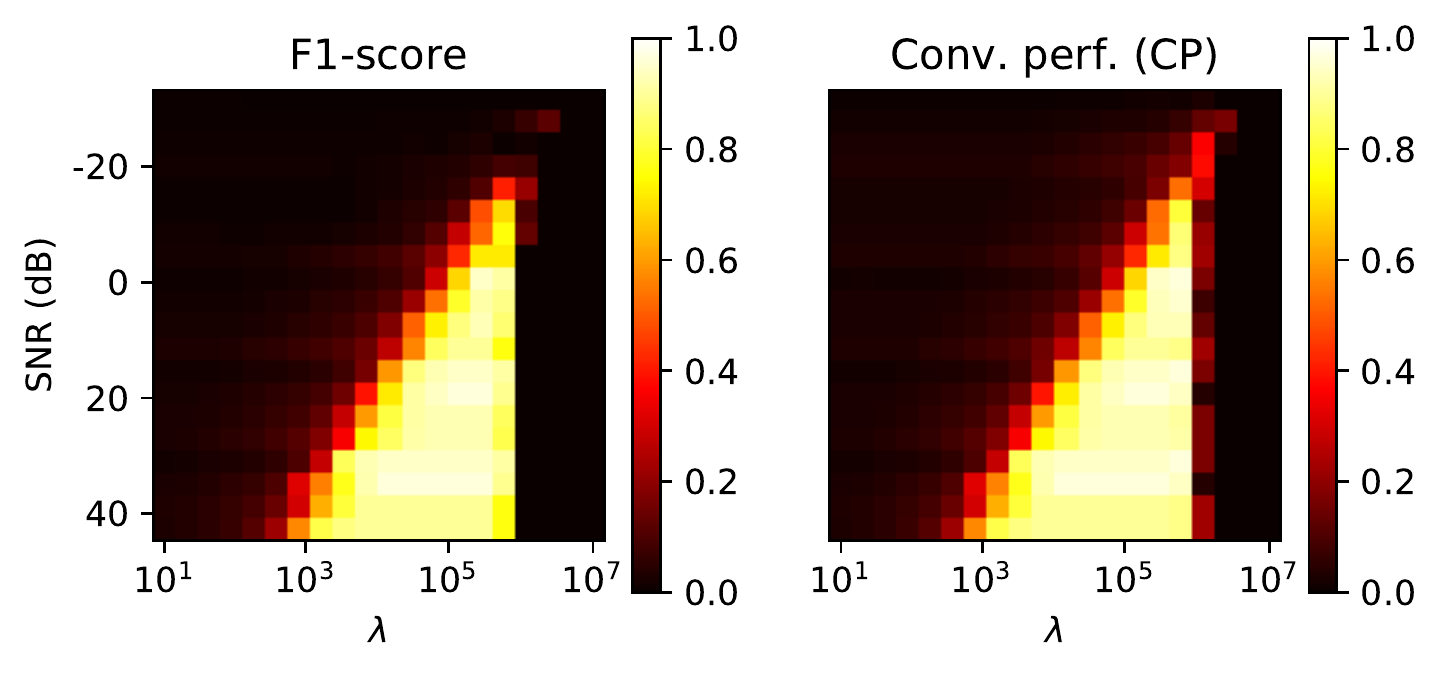}\vspace{-3mm}
		\caption{Influence of $\lambda$ and the signal-to-noise ratio on the performances of the Lasso estimator. Results are averaged over 5 draws.}
		\label{fig:noise_lambda_influence}
\end{figure}

\textbf{}
\label{sec:noise_lambda_influence}
{
Proper calibration of the regularization parameter $\lambda$ is crucial for the
success of the estimation. We want to visualize the influence of this choice,
especially for various noise levels. {Using real shapes of action potentials recorded in  \cite{bethus2012} and that have been already spike sorted by classical algorithms}, we simulate
signals of size $n=500$ for different noise levels, with $k=2$ neurons firing at
$50Hz$ and recorded by $d=4$ electrodes. A first measure of performance that we consider is the classical F1-score, which estimates a balance between false positive and false negative rates. This measure tends to be pessimistic as it penalizes equally short and long time deviations. We introduce a softer measure of performance: $ CP(x,y)= 1 - \norm{K*(x-y)}{1}/(\norm{x}{1} + \norm{y}{1})$, where $K$ is a normalized rectangular function. Depending on the size of the support of $K$, this measure allows us to tolerate small time deviations. 
}
The regularization parameter $\lambda$ should be chosen carefully in order for the Lasso to achieve good recovery. Taking $\lambda$ too small leads to too many activation whereas taking $\lambda$ too large leads to zero activation detected. As we can see on \cref{fig:noise_lambda_influence}, when the SNR is too weak, the admissible range for $\lambda$ becomes too narrow for any automatic tuning method to work. However for realistic signal-to-noise ratios, recovery is possible for a decent range of the regularization parameter $\lambda$, which confirms that the implementation of this method is conceivable in this spike sorting application.

\label{sec:comparison_with_other_ss}

\section{Conclusion}
\label{sec:conclusion}

{
In this paper, we provide an efficient algorithm to estimate the activations of the neurons. The windowed active set exhibits good performances, both in terms of estimation quality and execution time (scales linearly).

For future works, we intend to prove that the Lasso estimator recovers the correct support of activations, especially when the number of electrodes grows. Applications on real datasets will also be carried out, allowing us in particular to take into account the real structure of the noise. Finally, our windowed lasso open the door to online adaptation of the spikes shapes and we will investigate this by extending the works of \cite{mairal2009online}.
}

\bibliographystyle{IEEEtran}

\end{document}